\documentclass{article}

\usepackage[numbers]{natbib}
\usepackage[final]{neurips_2024}

\usepackage[utf8]{inputenc} %
\usepackage[T1]{fontenc}    %
\usepackage{hyperref}       %
\usepackage{url}            %
\usepackage{booktabs}       %
\usepackage{amsfonts}       %
\usepackage{nicefrac}       %
\usepackage{microtype}      %
\usepackage{xcolor}         %
\usepackage{tabularx}       
\usepackage{graphicx}
\usepackage{lipsum}
\usepackage{marvosym}
\usepackage{xspace}
\usepackage{blindtext}
\usepackage{amsmath}
\usepackage{bm}
\usepackage{algorithm}
\usepackage{algpseudocode}
\usepackage{wrapfig}
\usepackage{enumitem}
\usepackage{subfigure}
\usepackage{makecell}

\hypersetup{colorlinks,urlcolor=black,citecolor=cyan}

\makeatletter
\DeclareRobustCommand\onedot{\futurelet\@let@token\@onedot}
\def\@onedot{\ifx\@let@token.\else.\null\fi\xspace}

\def\eg{\emph{e.g}\onedot}

\makeatother

\definecolor{citecolor}{rgb}{0.21,0.49,0.74}
\definecolor{linkcolor}{HTML}{ED1C24}
\definecolor{graycolor}{rgb}{0.95,0.95,0.95}

\usepackage[capitalize]{cleveref}
\crefname{section}{Sec.}{Secs.}
\Crefname{section}{Section}{Sections}
\Crefname{table}{Table}{Tables}
\crefname{table}{Tab.}{Tabs.}
\crefname{figure}{Fig.}{Figs.}

\newcommand{\modelname}{HY-Motion 1.0}

\newcommand{\teamname}{Tencent Hunyuan 3D Digital Human Team}

\def\huggingface{\raisebox{-1.5pt}{\includegraphics[height=1.05em]{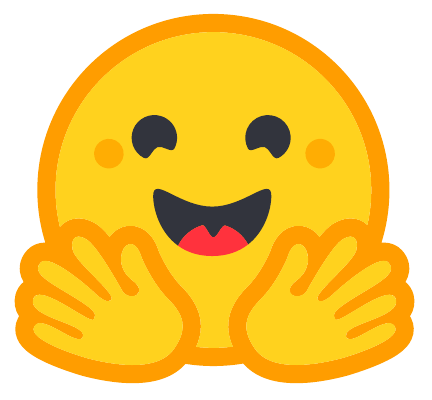}}}

\def\github{\raisebox{-1.5pt}{\includegraphics[height=1.05em]{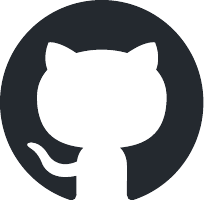}}}

\newcommand{\hflink}{https://huggingface.co/tencent/HY-Motion-1.0}
\newcommand{\ghlink}{https://github.com/Tencent-Hunyuan/HY-Motion-1.0}

\definecolor{ao}{rgb}{0.0, 0.0, 1.0}
\definecolor{airforceblue}{rgb}{0.36, 0.54, 0.66}
\definecolor{ceruleanblue}{rgb}{0.16, 0.32, 0.75}
\definecolor{cerulean}{rgb}{0.0, 0.48, 0.65}
\definecolor{celestialblue}{rgb}{0.29, 0.59, 0.82}
\definecolor{azure(colorwheel)}{rgb}{0.0, 0.5, 1.0}
\definecolor{babyblue}{rgb}{0.54, 0.81, 0.94}
\definecolor{babyblueeyes}{rgb}{0.63, 0.79, 0.95}
\definecolor{ballblue}{rgb}{0.13, 0.67, 0.8}

\definecolor{asparagus}{rgb}{0.53, 0.66, 0.42}
\definecolor{ao(english)}{rgb}{0.0, 0.5, 0.0}
\definecolor{applegreen}{rgb}{0.55, 0.71, 0.0}
\definecolor{armygreen}{rgb}{0.29, 0.33, 0.13}
\definecolor{gray-asparagus}{rgb}{0.27, 0.35, 0.27}
\definecolor{green(ryb)}{rgb}{0.4, 0.69, 0.2}

\definecolor{amethyst}{rgb}{0.6, 0.4, 0.8}
\definecolor{antiquefuchsia}{rgb}{0.57, 0.36, 0.51}
\definecolor{blue-violet}{rgb}{0.54, 0.17, 0.89}
\definecolor{brightlavender}{rgb}{0.75, 0.58, 0.89}
\definecolor{brightube}{rgb}{0.82, 0.62, 0.91}
\definecolor{brilliantlavender}{rgb}{0.96, 0.73, 1.0}

\definecolor{amber}{rgb}{1.0, 0.75, 0.0}
\definecolor{amber(sae/ece)}{rgb}{1.0, 0.49, 0.0}
\definecolor{atomictangerine}{rgb}{1.0, 0.6, 0.4}
\definecolor{burntorange}{rgb}{0.8, 0.33, 0.0}
\definecolor{burntsienna}{rgb}{0.91, 0.45, 0.32}
\definecolor{cadmiumorange}{rgb}{0.93, 0.53, 0.18}
\definecolor{carrotorange}{rgb}{0.93, 0.57, 0.13}
\definecolor{chocolate(web)}{rgb}{0.82, 0.41, 0.12}
\definecolor{chromeyellow}{rgb}{1.0, 0.65, 0.0}
\definecolor{darkorange}{rgb}{1.0, 0.55, 0.0}
\definecolor{darktangerine}{rgb}{1.0, 0.66, 0.07}
\definecolor{deepcarrotorange}{rgb}{0.91, 0.41, 0.17}
\definecolor{deepsaffron}{rgb}{1.0, 0.6, 0.2}
\definecolor{fulvous}{rgb}{0.86, 0.52, 0.0}

\title{
\modelname{}: Scaling Flow Matching Models for Text-To-Motion Generation
}

\author{
    \textbf{\teamname{}}
}

\begin{document}

\maketitle

\begin{abstract}
We present \textbf{HY-Motion 1.0}, a series of state-of-the-art, large-scale, motion generation models capable of generating 3D human motions from textual descriptions. 
HY-Motion 1.0 represents the first successful attempt to scale up Diffusion Transformer (DiT)-based flow matching models to the billion-parameter scale within the motion generation domain, delivering instruction-following capabilities that significantly outperform current open-source benchmarks. 
Uniquely, we introduce a comprehensive, full-stage training paradigm — including large-scale pretraining on over 3,000 hours of motion data, high-quality fine-tuning on 400 hours of curated data, and reinforcement learning from both human feedback and reward models — to ensure precise alignment with the text instruction and high motion quality. 
This framework is supported by our meticulous data processing pipeline, which performs rigorous motion cleaning and captioning. Consequently, our model achieves the most extensive coverage, spanning over 200 motion categories across 6 major classes. 
We release HY-Motion 1.0 to the open-source community to foster future research and accelerate the transition of 3D human motion generation models towards commercial maturity.
\end{abstract}

\begin{figure}[h]
\centering
\vspace{-15pt}
\includegraphics[width=0.9\linewidth]{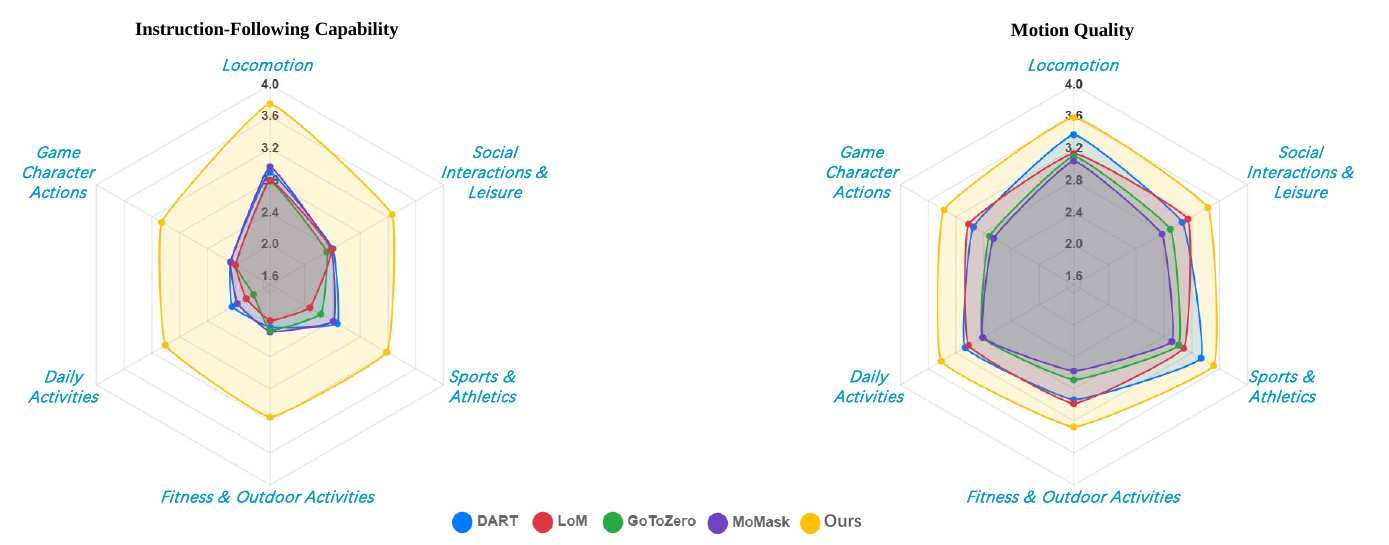}
\includegraphics[width=\linewidth]{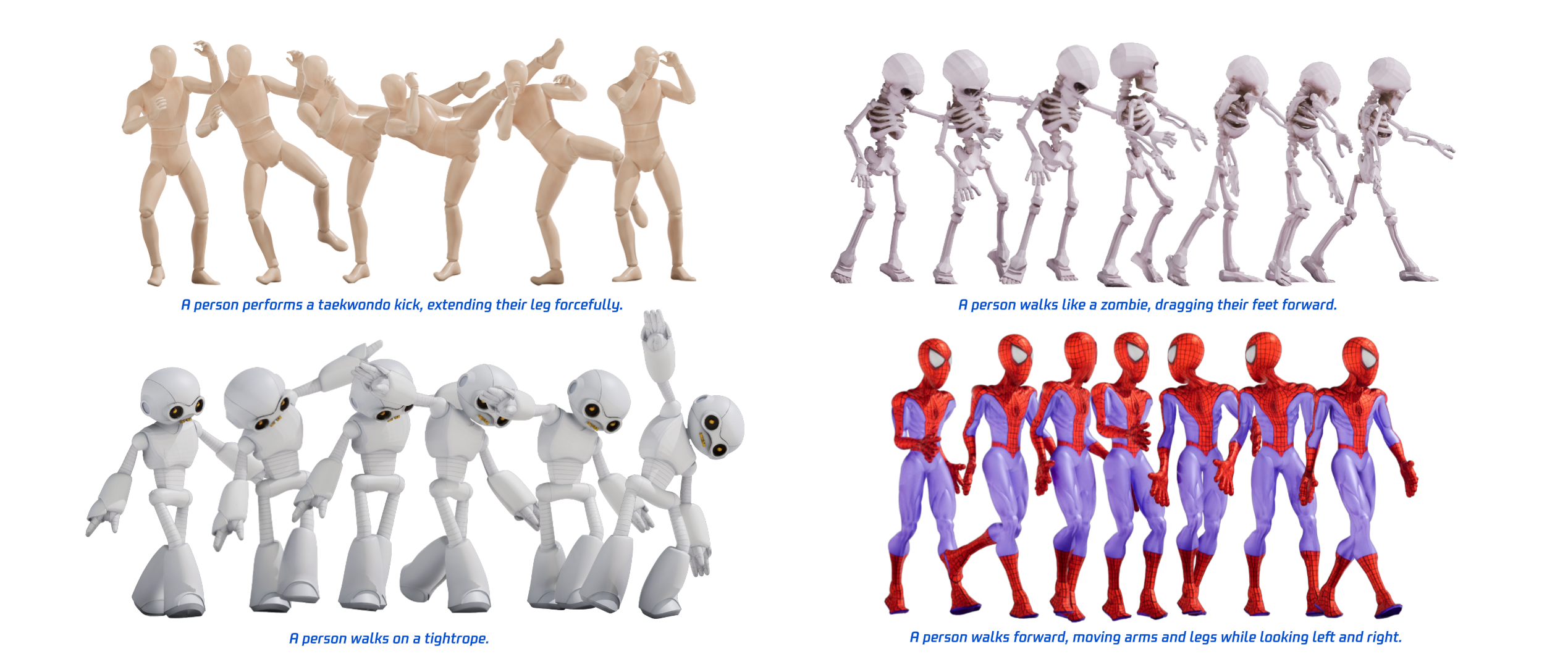}
\vspace{-20pt}
\caption{
    Top: Comparison of \modelname~to state-of-the-art text-to-motion models (DART \cite{DartControl}, LoM \cite{chen2024language}, GoToZero \cite{Fan_2025_ICCV}, and MoMask \cite{momask}). Bottom: Example results generated by \modelname~(retargeted to different characters). 
}
\label{fig:teasercomp}
\end{figure}

\section{Introduction}
\label{sec:intro}

The creation of high-quality 3D content has become a cornerstone of modern digital experiences, driving innovation in fields ranging from virtual and augmented reality (VR/AR) and interactive entertainment to robotics and digital human applications. 
Traditionally, generating realistic 3D human animations has required complex technical skills and a significant amount of time for well-trained animators, even with the assistance of expensive motion capture (MoCap) systems.
In recent years, the advance of generative AI has opened a new frontier: text-to-motion generation, a paradigm that aims to democratize 3D human animation by allowing creators to generate complex motions simply by describing them in natural language.

Significant progress in this domain has been fueled by various motion generation models \cite{mdm,momask,DartControl,motionlcm,remodiffuse,motiondiffuse,mld,Kim2025,Guo_2025_ICCV,Xiao_2025_ICCV,Pinyoanuntapong2025MaskControl} and motion datasets \cite{Guo2022,mahmood2019amass,Mason2022,Li2023,Kim2025,Harvey2020,Kobayashi2023,Guo2025}, with recent efforts exploring scaling motion generation with Large Language Models (LLMs) \cite{t2mgpt,motiongpt_nips,chen2024language,Lu_2025_CVPR,cao2025real,Fan_2025_ICCV}. 
On one hand, models such as MoMask \cite{momask} and DART \cite{DartControl} have shown promising results. However, being relatively small in model size, their generative capabilities are often limited, struggling to understand user instructions and produce motions of high complexity. 
On the other hand, models like LoM \cite{chen2024language} and GoToZero \cite{Fan_2025_ICCV}, have leveraged Large Language Models to generate motions by expanding vocabulary with discrete motion tokenizers. 
While this approach enhances semantic diversity, the quantization process inherent in discrete representations often leads to a degradation in motion quality, resulting in outputs that lack smoothness and naturalness. 
These divergent approaches highlight several persistent challenges that hinder the widespread adoption and commercial viability of text-to-motion models. First, motion fidelity and realism remain a primary concern; models frequently produce results with noticeable artifacts such as foot sliding, unnatural physics, or the aforementioned lack of smoothness. Second, the semantic alignment between the input text and the output motion remains a significant hurdle, as models often fail to accurately interpret complex or nuanced instructions. Third, unlike in image and video generation, the scaling of motion generation diffusion models has been relatively underexplored, leaving their full potential for enhanced quality and instruction-following largely untapped. Finally, all these challenges are compounded by a data bottleneck: the lack of large-scale, diverse, and meticulously cleaned and annotated motion datasets.

To address these critical gaps, we present HY-Motion 1.0, a new series of state-of-the-art, large-scale models for text-to-motion generation. 
Our work marks a significant leap forward by being the first to successfully scale a Diffusion Transformer (DiT)-based architecture \cite{DIT}, which leverages a flow matching objective \cite{Lipman2022,Esser2024,labs2025flux1kontextflowmatching}, to the billion-parameter scale in the motion generation domain. 
This scaling, combined with a novel and comprehensive training methodology, enables HY-Motion 1.0 to achieve unprecedented levels of motion quality and instruction-following precision, significantly outperforming existing open-source solutions.
To the best of our knowledge, we are the first to successfully scale a Diffusion Transformer (DiT)-based architecture \cite{DIT} to a billion-parameter scale, outperforming previous open-source state-of-the-art methods by a substantial margin, in terms of motion quality and the instruction-following capability.

Specifically, the foundation of our success lies in a full-stage training paradigm. 
Firstly, a large-scale pretraining is conducted on a motion dataset over 3,000 hours, including various motion categories and variations featuring maximal diversity and coverage, to grasp the motion structure and the semantic correlation between text and motion.
Secondly, we conduct fine-tuning on 400-hour high-quality text-motion pairs curated through meticulous motion filtering and caption filtering, resulting in better motion quality and more accurate instruction-following.
Crucially, we introduce a final alignment stage using reinforcement learning (RL) from both human feedback and reward models to ensure the generated motions are not only physically plausible but also precisely aligned with human preferences and textual commands. 
This entire framework is supported by our newly developed meticulous data processing pipeline, which integrates automatic motion data cleaning, processing, and captioning with extensive manual refinement, resulting in a dataset with the industry's widest coverage of over 200 distinct motion categories.

In this paper, we detail the data pipeline, model architecture, and training process of HY-Motion 1.0. Our primary contributions can be summarized as follows:

\begin{itemize}
    \item \textbf{Scaling Law for Text-to-Motion}: We are the first to successfully scale a DiT-based flow matching model to over one billion parameters for text-to-motion generation, demonstrating significant improvements in instruction following capabilities of text-to-motion models.
    \item \textbf{Comprehensive Full-Stage Training Paradigm}: We implement a three-stage training framework for the text-to-motion model training — large-scale pretraining, high-quality fine-tuning, and reinforcement learning — that holistically enhances model performance.
    \item \textbf{Meticulous Data Curation Pipeline}: We designed and implemented a meticulous data curation pipeline, including both automated processing and extensive manual refinement for data cleaning and captioning. This rigorous process has yielded a large-scale, high-quality motion dataset, 
    enabling the training of a large-scale network capable of generating high-quality motions accurately following the text instruction.
    \item \textbf{Open-Source Release}: We are releasing the HY-Motion 1.0 models to the open-source community to foster further research and accelerate the development of commercially viable 3D motion generation technologies.
\end{itemize}

\section{Data}
\label{sec:data}

\begin{figure}[h!]
    \centering
    \includegraphics[width=\linewidth, trim=30 150 100 150, clip]{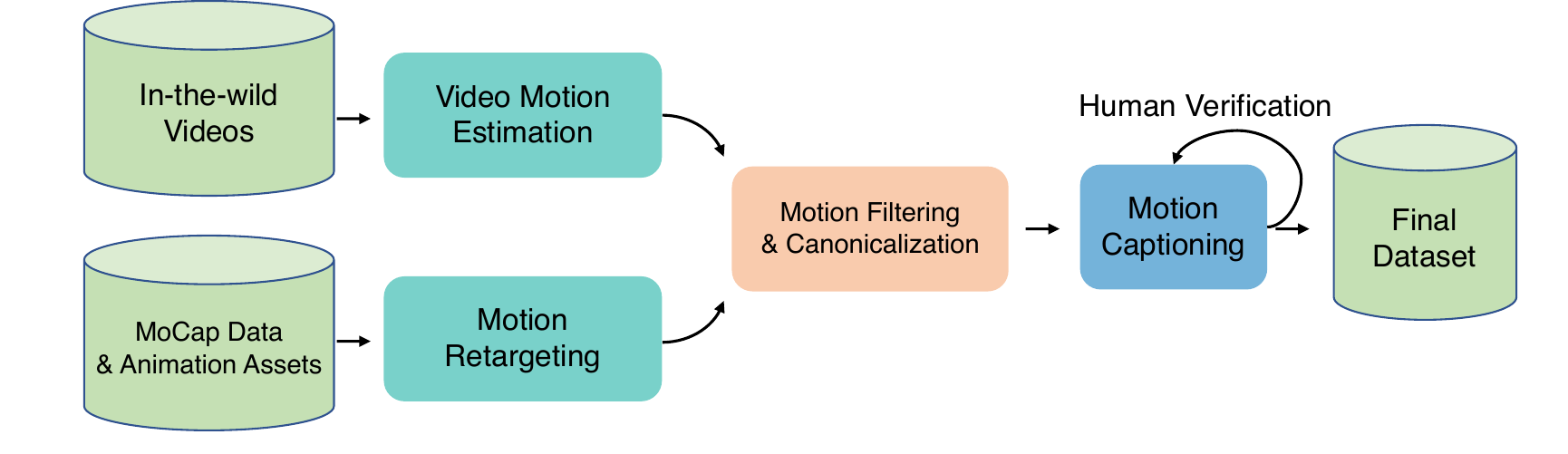}
    \vspace{-10pt}
    \caption{Overview of the data processing pipeline.}
    \label{fig:data_pipeline}
\end{figure}

Our data curation process consisted of acquisition, processing, filtering, and captioning, as illustrated in Fig. \ref{fig:data_pipeline}. We first describe our data acquisition, processing, and filtering in \cref{subsec:data_source}. Then we present our motion captioning process in \cref{subsec:captioning} and motion taxonomy in \cref{subsec:stats}.

\subsection{Data Acquisition and Filtering}
\label{subsec:data_source}

\paragraph{Data Acquisition.}

Our dataset was constructed from three complementary data sources: in-the-wild human motion videos, motion capture data, and 3D animation assets.
Human motion videos provided rich and diverse actions captured across varied real-world scenarios.
Motion capture data, typically recorded in controlled indoor environments, offered high-quality motion data but were limited in scene diversity.
3D animation assets, hand-crafted by professional artists for game production, exhibited exceptional motion quality but were relatively expensive and limited in quantity.

For video data, we started with a massive collection of 12 million high-quality, in-the-wild video clips from HunyuanVideo~\citep{hunyuanvideo2025}.
We first processed these clips through a rigorous pre-processing stage, employing shot boundary detection to segment videos into coherent scenes and a human detector to search for clips containing human subjects.
Subsequently, for the resulting candidate clips, we utilized a state-of-the-art human motion extraction algorithm GVHMR~\citep{shen2024gvhmr}, to reconstruct 3D human tracks of SMPL-X~\citep{smplx} parameters.
For motion capture data and 3D animation assets, we acquired approximately 500 hours of motion sequences.

\paragraph{Motion Processing and Filtering}

To facilitate downstream model training, we standardized all motion data onto a unified SMPL-H skeleton. We processed all motions in three steps: retargeting, low-quality motion filtering, and canonicalization.
We retargeted all motions onto a neutral SMPL-H skeleton~\citep{smplh}.
For motion data originally in SMPL~\citep{smpl}, SMPL-H~\citep{smplh}, or SMPL-X~\citep{smplx} formats, we employed mesh fitting to convert them into the unified SMPL-H representation.
For data with other skeletal structures, we applied a retargeting tool to map them to the SMPL-H skeleton.
For all retargeted data, we established a comprehensive filtering pipeline to remove low-quality motion clips.
This included removal of duplicates, abnormal poses, outliers of joint velocity, detection of anomalous displacements, pruning of static motions, and detection of artifacts such as foot-sliding.
After filtering, we applied canonicalization to standardize the data.
All motions were resampled to 30 fps, and sequences longer than 12 seconds were segmented into multiple clips.
Each motion was normalized to a canonical coordinate frame: Y-axis up, the starting position centered at the origin, the lowest body point aligned on the ground plane, and initial facing direction along the positive Z-axis.
Ultimately, we obtained over 3000 hours of motion data, including 400 hours of high-quality 3D motion data.

\subsection{Motion Captioning}
\label{subsec:captioning}

Our captioning pipeline primarily leveraged vision-language models (VLMs) for automatic annotation, followed by manual verification and refinement. Finally, we employed large language models (LLMs) to diversify and enrich the textual descriptions.

For video-sourced data, we directly used the original videos corresponding to each motion sequence.
For 3D motion data, we textured and rendered the SMPL-H model to generate synthetic videos.
These videos, along with optimized prompts dedicated to human motion, were fed into a VLM (\eg, Gemini-2.5-Pro) to obtain preliminary captions and action keywords.
To obtain as accurate a text-motion pair as possible for the high-quality data (i.e., rendered motions),
we conducted manual verification on the VLM outputs, correcting erroneous descriptions and supplementing missing key motion components.

For all generated captions, we utilized an LLM to standardize the caption structure while preserving the original semantics and to create diverse paraphrases for data augmentation.

\subsection{Taxonomy}
\label{subsec:stats}

\begin{figure*}[t]
    \centering
    \begin{minipage}[b]{\textwidth}
        \centering
        \includegraphics[width=\textwidth, trim=0 20 0 20]{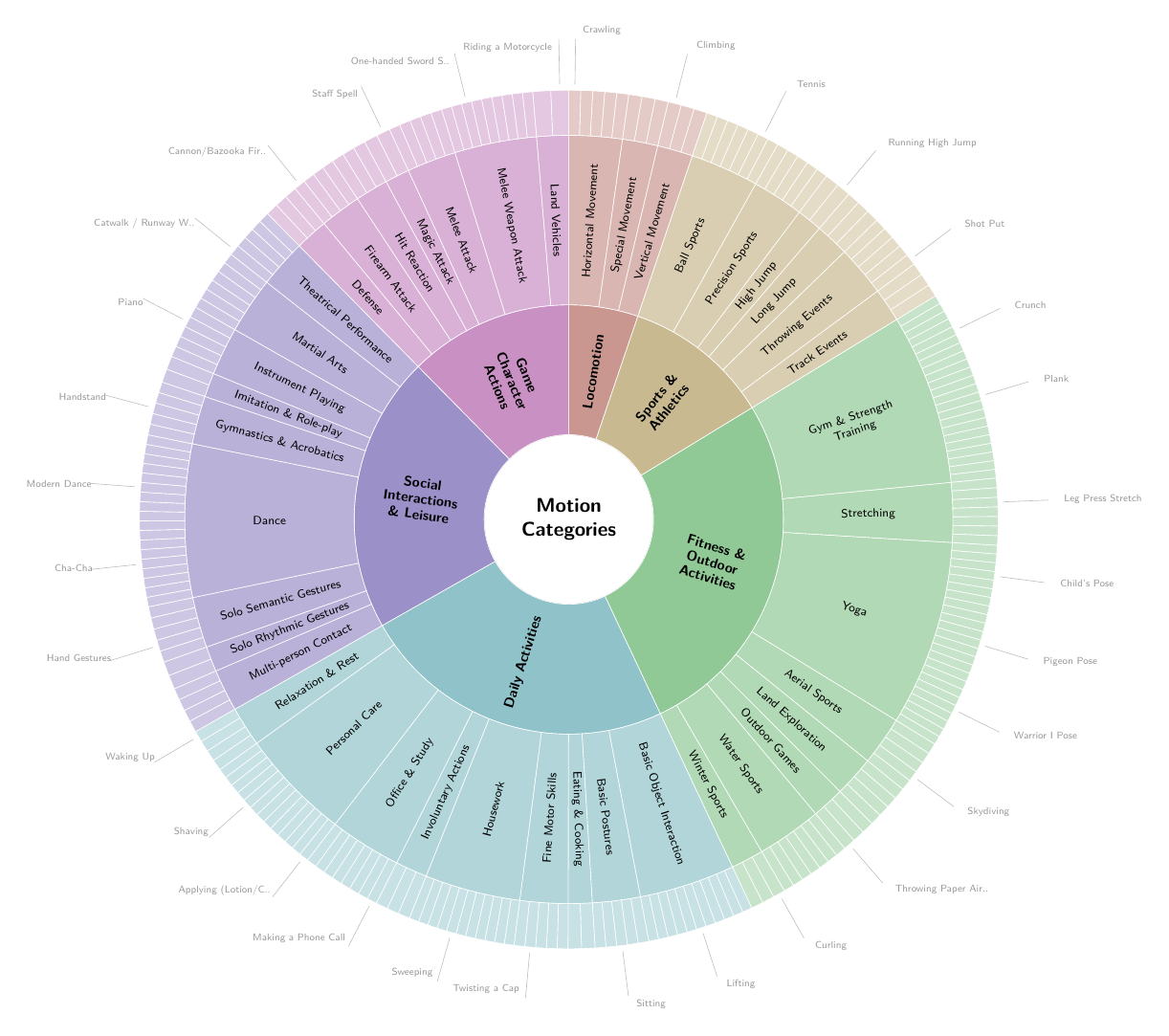}
    \end{minipage}
    \hfill
    \caption{The hierarchy of our motion categories.}
    \label{fig:data_overview}
\end{figure*}

To organize our dataset, we established a three-level hierarchical motion taxonomy based on textual captions and keywords. As illustrated in Figure \ref{fig:data_overview}, the hierarchy began with six coarse-grained, top-level categories: (a) Locomotion, (b) Sports \& Athletics, (c) Fitness \& Outdoor Activities, (d) Daily Activities, (e) Social Interactions \& Leisure, and (f) Game Character Actions. These categories were progressively subdivided, culminating in over 200 fine-grained motion classes at the leaf level.

\section{Model Design}
\label{sec:method}

\begin{figure*}[h]
    \centering
    \includegraphics[width=\linewidth, trim=10 140 40 120, clip]{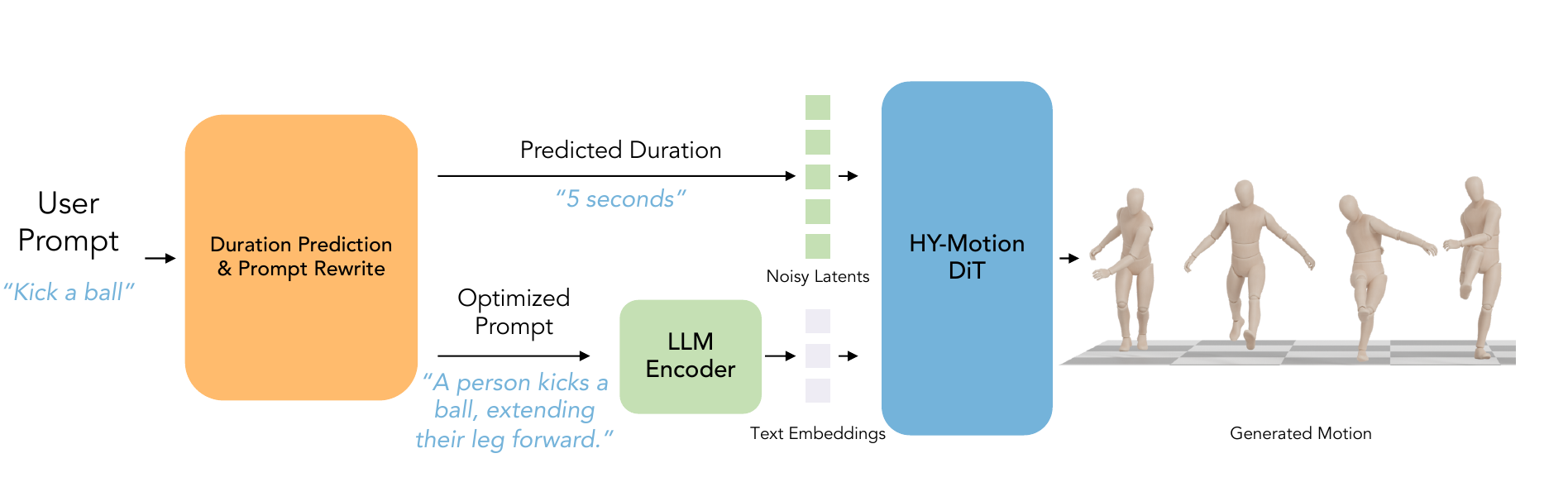}
    \vspace{-10pt}
    \caption{
        Overview of the \modelname{} framework.
    }
    \label{fig:pipeline}
\end{figure*}

The overview framework of \modelname{} is shown in Fig. \ref{fig:pipeline}.
The core is our HY-Motion DiT model, which accepts a text prompt and an expected duration of the motion as inputs and generates a clip of 3D human motion as outputs.
We first detail our motion representation in \cref{subsec:motion_rep}.
Then the HY-Motion DiT model is described in \cref{subsec:backbone}.
Besides the core model, we train an independent LLM (\cref{subsec:pipeline_enhancement}) to analyze user prompts and predict the expected duration of the motion clips, and additionally, convert user prompts into better-structured prompts.

\subsection{Motion Representation}
\label{subsec:motion_rep}

We employ the skeleton definition of SMPL-H~\citep{Romero2017} (22 joints without hands).
Formally, we denote a motion sequence as $\bm{x}=\{\bm{f}_1,\bm{f}_2,\dots,\bm{f}_N\}$, where each frame is represented as a vector $\bm{f}\in\mathbb{R}^{201}$, comprising the global root translation $\bm{t}\in\mathbb{R}^3$, the global body orientation $\bm{r}\in\mathbb{R}^6$, the local joint rotations $\bm{j}^r\in\mathbb{R}^{21\times6}$, and the local joint positions $\bm{j}^p\in\mathbb{R}^{22\times3}$.
All rotational parameters adhere to the continuous 6D representation~\citep{Zhou2019}.
Noticeably, our representation is similar to DART~\citep{DartControl}, where the rotation representation is compatible with typical animation workflows, which differs from the commonly used HumanML3D~\citep{Guo2022} representation.
We further remove the explicit temporal derivatives (velocities) and foot contact labels, as we observed faster training convergence.

\subsection{HY-Motion DiT}
\label{subsec:backbone}

\begin{figure*}[h]
    \centering
    \includegraphics[width=\linewidth]{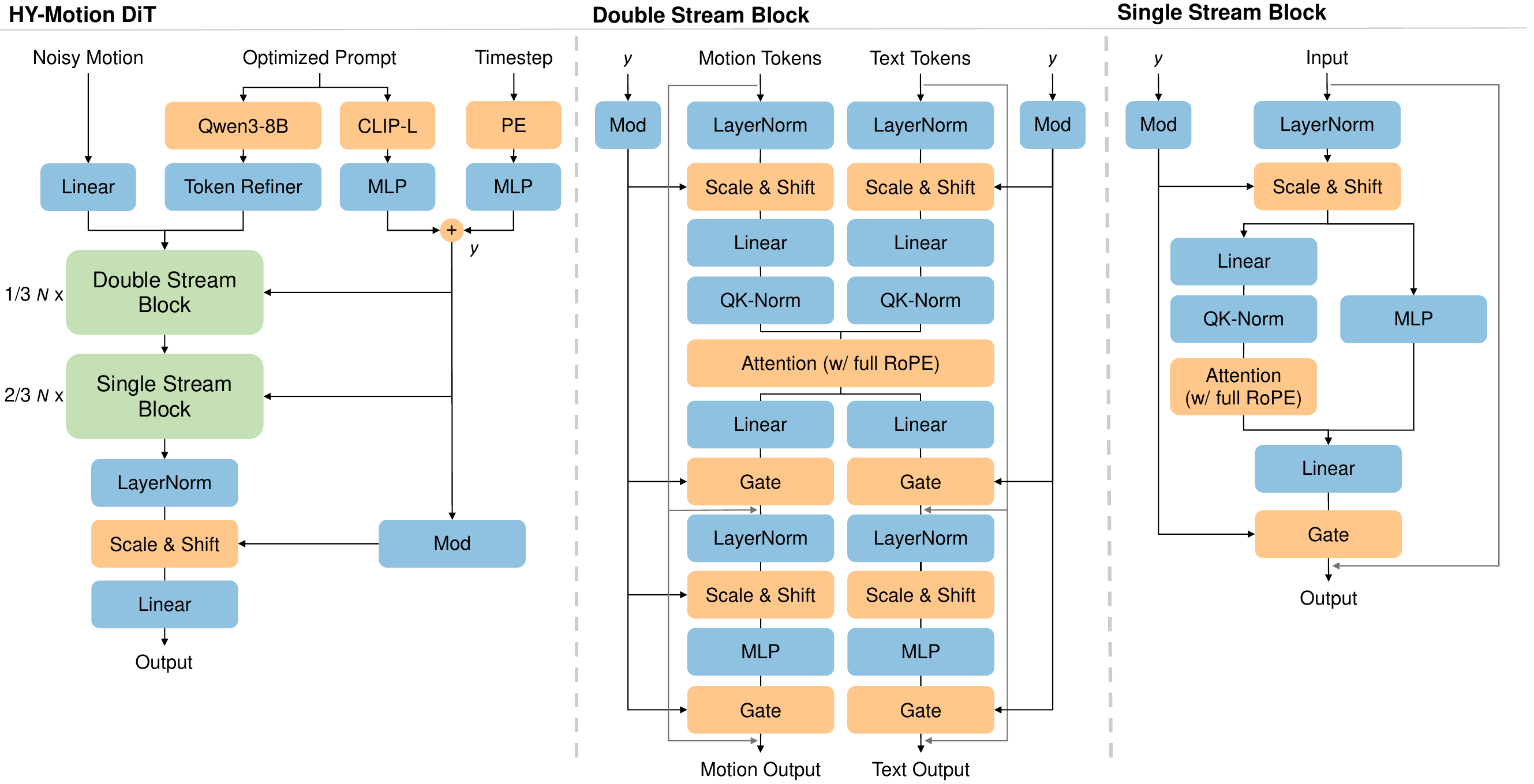}
    \vspace{-10pt}
    \caption{
        Model architecture of our HY-Motion DiT.
    }
    \label{fig:backbone_mmdit}
\end{figure*}

\paragraph{Model Architecture.}
Similar to HunyuanVideo~\cite{Kong2024}, we employ a hybrid Transformer that combines dual-stream and single-stream processing to model the joint distribution of motion and text.
The network begins with dual-stream blocks, where motion latents and text tokens are processed via independent QKV projections and MLPs.
Crucially, they interact through a joint attention mechanism, allowing motion features to query semantic cues from the text while preserving their distinct, modality-specific representations.
These streams subsequently merge in the single-stream blocks, where motion and text tokens are concatenated into a unified sequence.
Here, we employ parallel spatial and channel attention modules to facilitate deep multimodal fusion and information exchange.

\paragraph{Text Encoders.}
We leverage a hierarchical dual-conditioning strategy to integrate both fine-grained and global text guidance.
Firstly, we employ Qwen3-8B~\citep{Yang2025} to extract rich, token-wise semantic embeddings.
Addressing the limitation that LLMs typically utilize causal attention—which restricts the context for non-autoregressive generation, we follow HunyuanVideo~\citep{Kong2024} by adopting a Bidirectional Token Refiner.
This module transforms the causal LLM features into bidirectional representations before injecting them into the dual-stream blocks, thereby providing holistically contextualized guidance for the diffusion model.
Complementarily, we utilize CLIP-L~\citep{Radford2021} to extract a global text embedding.
This embedding is concatenated with the timestep embedding and injected via a separated AdaLN~\citep{Peebles2023} mechanism, where layer-specific modulation parameters are learned to adaptively regulate feature statistics throughout the network.

\paragraph{Attention Mechanism and Positional Encoding.}
We devise a composite Attention Mechanism that incorporates specific masking strategies for both cross-modal interaction and temporal modeling.
First, to regulate multimodal information flow, we enforce an asymmetric attention mask.
While motion tokens attend globally to the text sequence to extract semantic cues, text tokens are explicitly masked from the motion latents.
This unidirectional constraint prevents the diffusion noise inherent in the motion states from propagating back to the text embeddings, thereby preserving the integrity of the semantic conditioning.
Second, for temporal modeling within the motion branch, we implement a narrow band mask strategy.
Predicated on the hypothesis that kinematic dynamics are governed primarily by local continuity, we restrict attention to a sliding window of $121$ frames under $30$ fps.
This imposes a locality inductive bias that decomposes long sequences into coherent substructures, effectively focusing the model on complex local dynamics while ensuring linear computational complexity.
To spatially ground these attention interactions, we optimize the Positional Encoding by adopting full Rotary Positional Embeddings (RoPE) \cite{su2024roformer}.
Diverging from disjoint encoding schemes, we concatenate text and motion embeddings into a single sequence prior to encoding.
By applying RoPE across this unified sequence, we establish a continuous relative coordinate system, enabling the model to inherently resolve the positional correspondence between specific textual tokens and temporal motion frames.

\paragraph{Flow Matching Objective.}
We employ Flow Matching~\citep{Lipman2022} to construct a continuous probability path that bridges the standard Gaussian noise distribution and the complex motion data distribution.
We adopt the optimal transport path, defined as a linear interpolation $\bm{x}_t=(1-t)\bm{x}_0 + t\bm{x}_1$, which implies a constant target velocity~\citep{Liu2022}.
The training objective is to minimize the Mean Squared Error (MSE) between the predicted and ground-truth velocity:
\begin{equation}
    \mathcal{L}_{\text{FM}} = \mathbb{E}_{t, \bm{x}_0, \bm{x}_1} [ || \bm{v}_{\bm{\theta}}(\bm{x}_t,\bm{c},t) - \bm{v}_t ||^2_2 ],
    \label{eq:flow_matching}
\end{equation}
where $\bm{x}_1$ represents the clean motion data, $\bm{x}_0\sim\mathcal{N}(0, \bm{I})$ is the initial noise, and the target velocity is given by $\bm{v}_t=\bm{x}_1-\bm{x}_0$.
During inference, the generation process is formulated as an Ordinary Differential Equation (ODE): $d\bm{x}/d\bm{t}=\bm{v}_{\bm{\theta}}(\bm{x}_t,\bm{c},t)$.
Starting from random noise $\bm{x}_0$, we recover the clean motion $\bm{x}_1$ by numerically integrating this ODE along the predicted velocity field using an ODE solver (\eg, Euler).

\subsection{Duration Prediction \& Prompt Rewrite}
\label{subsec:pipeline_enhancement}

To accommodate diverse user inputs, we employ a dedicated LLM for both duration prediction and prompt rewrite. This module first predicts the target motion's duration from the user prompt. It then transforms the prompt into a structured format optimized for our DiT model. This design is motivated by the observation that LLMs possess inherent common-sense knowledge about the typical duration of human activities, enabling them to infer temporal length from textual descriptions. To further enhance its accuracy, we fine-tune the LLM on a dataset of ground-truth motion durations, which aligns its predictions with our training data distribution. Crucially, this fine-tuning preserves the model's ability to generalize to unseen motion descriptions.

\paragraph{Data Synthesis.}
Our dataset for fine-tuning the LLM consists of data triplets \{\textit{user prompt}, \textit{optimized prompt}, \textit{duration}\}. The ``optimized prompt'' and ``duration'' are the ground-truth text caption and duration of each motion clip in our motion dataset, respectively.
To simulate realistic ``user prompt'', we employed a powerful LLM (\eg, Gemini-2.5-Pro) to synthesize a dataset of user queries. Using the ground-truth text captions as a basis, we prompted the LLM to generate inputs that mimic the nature of casual human commands. These synthetic queries are intentionally diverse, encompassing informal language, a mix of English and Chinese, and varying levels of specificity, including queries that are deliberately brief and ambiguous.

\paragraph{Two-Stage Fine-Tuning.}
The duration prediction \& prompt rewrite module is fine-tuned from a Qwen3-30B-A3B~\citep{Yang2025} model with a two-stage training procedure.
\begin{itemize}
    \item \textit{Supervised Fine-Tuning (SFT):}
    The model is first fine-tuned on the constructed dataset with ground-truth supervision.
    It learns to sequentially generate ``optimized prompt'' and ``duration'' from the ``user prompt'', effectively grounding the temporal prediction in the semantic context of the rewritten text.

    \item \textit{Reinforcement Learning (RL):}
    To further mitigate hallucinations and enhance fidelity, we reinforce the model using Group Relative Policy Optimization (GRPO)~\citep{Shao2024}.
    We utilize a more powerful model (Qwen3-235B-A22B-Instruct-2507~\citep{Yang2025}) as a reward judge to evaluate multiple candidate outputs for each prompt.
    The reward function assesses two critical dimensions: semantic consistency, ensuring the rewrite remains faithful to the user's intent, and temporal plausibility, verifying that the predicted duration aligns with the physical complexity of the described action.
    By optimizing against the relative advantages of these candidates, GRPO steers the policy toward generating instructions that are both semantically precise and temporally coherent.
\end{itemize}

\section{Model Training}
\label{sec:training}

The training process for HY-Motion DiT consists of three stages: large-scale pretraining, high-quality fine-tuning, and reinforcement learning. 
To reconcile the trade-off between generalization (requiring massive scale) and precision (requiring high-quality data), we implement a ``scale-then-refine'' curriculum for the supervised training phases (\cref{subsec:stage1} and \cref{subsec:stage2}).
We begin with large-scale pretraining on an expansive dataset comprising over $3,000$ hours of motion data, enabling the model to assimilate a broad motion prior and learn a generalized representation of human movement.
This is followed by a high-quality fine-tuning stage on nearly $400$ hours of meticulously curated and annotated data, which serves to sharpen the model's capabilities and refine the details of the generation.
While these supervised stages effectively model the underlying data distribution, strictly imitating the training distribution does not necessarily maximize perceptual quality or semantic responsiveness.
Therefore, we introduce a final reinforcement learning phase (\cref{subsec:stage3}) to bridge the gap between ``statistical likelihood'' and ``human preference''.
By incorporating feedback from reward models, this stage fine-tunes the model to optimize for perceptual motion quality and maximize adherence to complex user instructions—metrics that are often difficult to capture through supervised loss functions alone.

\subsection{Large-Scale Pretraining}
\label{subsec:stage1}

The primary objective of this foundational stage is general motion prior acquisition.
To maximize the model's exposure to diverse semantic scenarios and kinematic patterns, we train the model on our full motion dataset $\mathcal{D}_{\text{all}}$ consisting of over $3,000$ hours of motion data. 
This dataset prioritizes coverage over precision, comprising both high-quality captures and noisy, in-the-wild video extractions, accompanied by text descriptions ranging from manual annotations to VLM-generated annotations.
In this phase, we employ a standard Flow Matching objective (\cref{eq:flow_matching}) with a constant learning rate $\eta_{\text{pre}}$, aiming to rapidly establish a broad support for the motion distribution.

From a distributional perspective, this stage allows the model to effectively ``learn to move''.
We observe a rapid convergence in training loss, indicating that the model successfully captures the fundamental dynamics of human movement and establishes a robust mapping between diverse textual prompts and plausible poses.
The model demonstrates strong semantic generalization, capable of responding to a vast vocabulary of actions.
However, the generated motion quality inevitably mirrors the mixed nature of the input distribution.
Since a significant portion of $\mathcal{D}_{\text{all}}$ contains noisy data, the model's output distribution remains high-entropy and loosely constrained.
Consequently, while semantically correct, the generated sequences often exhibit artifacts such as high-frequency jitter, foot sliding, and minor anatomical inconsistencies.
We treat this result as an expected outcome: the model has successfully learned a generalized but coarse prior, providing a semantic-rich initialization that is ready to be "sharpened" in the subsequent fine-tuning stage.

\subsection{High-Quality Fine-Tuning}
\label{subsec:stage2}

Following the broad acquisition of motion priors, the second stage focuses on distribution refinement.
To bridge the gap between ``plausible'' and ``high-quality'', we transition the training source to the meticulously curated subset $\mathcal{D}_{\text{HQ}}$.
This dataset is characterized by high-quality, rigorously filtered motion data, paired with manually corrected textual descriptions.
The objective here is to concentrate the model's probability density around the optimal modes of the motion manifold.
Crucially, to refine fine-grained kinematic details while preventing the forgetting of the broad semantic priors learned in \cref{subsec:stage1}, we decay the learning rate to $\eta_{\text{ft}} = 0.1 \times \eta_{\text{pre}}$. 
This conservative optimization strategy ensures that the model polishes its output distribution without collapsing its generative diversity.

Empirically, this phase drives a qualitative evolution from ``roughly correct'' to ``precise and smooth''.
Kinematically, we observe that the model effectively suppresses the noise patterns inherited from the pretraining phase, drastically reducing high-frequency jitter and foot sliding artifacts while enforcing stricter anatomical consistency.
Semantically, the alignment becomes significantly more rigorous; the model demonstrates enhanced sensitivity to fine-grained anatomical instructions—such as accurately distinguishing between ``waving the left hand'' and ``waving the right hand''—which were often ambiguous in the noisy pretraining data.
Importantly, this sharpening of quality and control is achieved without a significant degradation in motion diversity, validating the efficacy of our coarse-to-fine curriculum.

\subsection{Reinforcement Learning}
\label{subsec:stage3}

While supervised learning establishes a strong kinematic prior, optimizing for data likelihood does not strictly equate to maximizing perceptual fidelity or instruction adherence.
Generative models often suffer from a misalignment gap where statistically probable motions may still exhibit physical artifacts or semantic ambiguity.
To bridge this gap, we propose a synergistic two-phase reinforcement learning, progressing from human preference learning to explicit objective constraint satisfaction.

\paragraph{Human Preference Alignment via DPO.}
We first aim to internalize the nuanced standards of human perception.
Since defining a closed-form reward function for complex semantic alignment and motion quality is intractable, we leverage Direct Preference Optimization (DPO)~\citep{Rafailov2023} to steer the policy directly from human feedback.
We utilize the Stage-2 model to generate candidate pairs for a diverse prompt set.
From an annotated pool of $40,000$ pairs, we curate a high-information subset of $9,228$ pairs where human judges identified a distinct ``winner''($\bm{x}_w$) and ``loser''($\bm{x}_l$) based on instruction adherence and visual plausibility.
By maximizing the likelihood margin between these pairs, DPO implicitly models the latent reward function of human perception:
\begin{equation}
    \mathcal{L}_{\text{DPO}}(\pi_{\bm{\theta}};\pi_{\text{ref}}) = - \mathbb{E}_{(\bm{c}, \bm{x}_w, \bm{x}_l) \sim \mathcal{D}_{\text{pref}}} \left[ \log \sigma \left( \beta \log \frac{\pi_{\bm{\theta}}(\bm{x}_w|\bm{c})}{\pi_{\text{ref}}(\bm{x}_w|\bm{c})} - \beta \log \frac{\pi_{\bm{\theta}}(\bm{x}_l|\bm{c})}{\pi_{\text{ref}}(\bm{x}_l|\bm{c})} \right) \right],
\end{equation}
where $\pi_{\text{ref}}$ represents the frozen reference policy, and $\mathcal{D}_{\text{pref}}=\{(\bm{c}^i, \bm{x}_w^i, \bm{x}_l^i)\}_{i=1}^N$.
This phase effectively acts as a semantic anchor, drastically improving the model's ``Pass Rate'' by pruning low-quality modes that, while kinetically valid, fail to align with user intent.

\paragraph{Physics \& Semantic Refinement via Flow-GRPO.}
Building upon the preference-aligned policy, we address the limitations of DPO in enforcing strict physical and semantic boundaries.
We employ Flow-GRPO~\citep{Liu2025}, a variant of Group Relative Policy Optimization tailored for flow matching models, to optimize explicit objectives.
Unlike standard PPO which requires a value network, Flow-GRPO stabilizes training by normalizing advantages within a group of sampled outputs for a given prompt.
We maximize the following objective:
\begin{equation}
    \mathcal{L}_{\text{Flow-GRPO}}(\pi_{\bm{\theta}};\pi_{\text{ref}}) = \mathbb{E}_{\bm{c} \sim \mathcal{D}_\text{GRPO}, \{\bm{x}^i\}_{i=1}^G \sim \pi_{\text{ref}}(\cdot| \bm{c})} \left[ f(r,\hat{A},\bm{\theta}, \epsilon,\beta) \right],
\end{equation}
where the objective function $f$ is averaged over time steps $T$:
\begin{equation}
    \begin{aligned}
        f(r,\hat{A},\bm{\theta}, \epsilon,\beta) &= \frac{1}{G}\sum_{i=1}^G \frac{1}{T} \sum_{t=1}^T \Big( \min \big( r_t^i(\bm{\theta})\hat{A}^i, \text{clip}(r_t^i(\bm{\theta}), 1-\epsilon, 1+\epsilon)\hat{A}^i \big) \\
        & \qquad - \beta D_{\text{KL}}(\pi_{\bm{\theta}} \| \pi_{\text{ref}}) \Big), \
        \text{with}\quad r_t^i(\bm{\theta}) = \frac{p_{\bm{\theta}}(\bm{x}_{t-1}^i|\bm{x}_{t}^i,\bm{c})}{p_{\text{ref}}(\bm{x}_{t-1}^i|\bm{x}_{t}^i,\bm{c})}.
    \end{aligned}
\end{equation}
Here, $\mathcal{D}_\text{GRPO}$ denotes the prompt dataset used for GRPO, $G$ denotes the group size, and $\hat{A}^i$ is the standardized advantage derived from the composite reward $R(\bm{x}^i, \bm{c})$.
Specifically, the reward function integrates two critical metrics:
a semantic reward ($R_{\text{sem}}$)  evaluated by a custom-trained Text-Motion Retrieval (TMR) model \cite{petrovich2023tmr}, and a physical reward ($R_{\text{phy}}$) that imposes hard penalties on artifacts such as foot sliding and root drift.
By explicitly optimizing this objective, the model is fine-tuned to satisfy rigorous kinematic constraints while maximizing semantic precision.

\section{Evaluation}
\label{sec:eval}

\subsection{Comparison with State-of-the-Art}
\label{subsec:main_results}

\begin{table*}[tb]
    \centering
    \caption{\textbf{Instruction-following capability comparison} with state-of-the-art text-to-motion models. Motion categories: (a) Locomotion, (b) Sports \& Athletics, (c) Fitness \& Outdoor Activities, (d) Daily Activities, (e) Social Interactions \& Leisure, and (f) Game Character Actions.}
    \vspace{5pt}
    \label{tab:sota_comparison}
    \begin{tabular}{l|cccccc|c|c}
        \toprule
        Method & Catg. (a) & (b) & (c) & (d) & (e) & (f) & Avg. & SSAE  \\
        \midrule
        MoMask \cite{momask} & 2.98 & 2.41 & 2.09 & 2.07 & 2.38 & 1.97 & 2.31 & 58.0\% \\
        GoToZero \cite{Fan_2025_ICCV} & 2.80 & 2.23 & 2.07 & 2.00 & 2.32 & 1.74 & 2.19 & 52.7\% \\
        DART \cite{DartControl} & 2.91 & 2.47 & 2.03 & 2.07 & 2.40 & 2.05 & 2.31  & 42.7\% \\
        LoM \cite{chen2024language} & 2.81 & 2.07 & 1.95 & 2.00 & 2.39 & 1.84 & 2.17  & 48.9\% \\
        \midrule
        \textbf{\modelname{}} & \textbf{3.76} & \textbf{3.18} & \textbf{3.15} & \textbf{3.06} & \textbf{3.25} & \textbf{3.01} & \textbf{3.24} & \textbf{78.6\%} \\
        \bottomrule
    \end{tabular}
    \vspace{-10pt}
\end{table*}

\begin{table*}[tb]
    \centering
    \caption{\textbf{Motion quality comparison} with state-of-the-art text-to-motion models. Motion categories: (a) Locomotion, (b) Sports \& Athletics, (c) Fitness \& Outdoor Activities, (d) Daily Activities, (e) Social Interactions \& Leisure, and (f) Game Character Actions.}
    \vspace{5pt}
    \label{tab:sota_comparison_mq}
    \begin{tabular}{l|cccccc|c}
        \toprule
        Method & Catg. (a) & (b) & (c) & (d) & (e) & (f) & Avg.  \\
        \midrule
        MoMask \cite{momask} & 3.05 & 2.91 & 2.58 & 2.66 & 2.77 & 2.81 & 2.79 \\
        GoToZero \cite{Fan_2025_ICCV} & 3.11 & 3.01 & 2.69 & 2.72 & 2.89 & 2.81 & 2.86  \\
        DART \cite{DartControl} & 3.38 & 3.33 & 2.94 & 2.95 & 3.06 & 3.07 & 3.11  \\
        LoM \cite{chen2024language} & 3.14 & 3.08 & 2.98 & 3.01 & 3.14 & 3.01 & 3.06  \\
        \midrule
        \textbf{\modelname{}} & \textbf{3.59} & \textbf{3.51} & \textbf{3.28} & \textbf{3.37} & \textbf{3.43} & \textbf{3.41} & \textbf{3.43} \\
        \bottomrule
    \end{tabular}
\end{table*}

\begin{figure}[tb]
    \centering
    \begin{tabular}{c}
        \includegraphics[width=0.9\linewidth, trim=50 450 45 30]{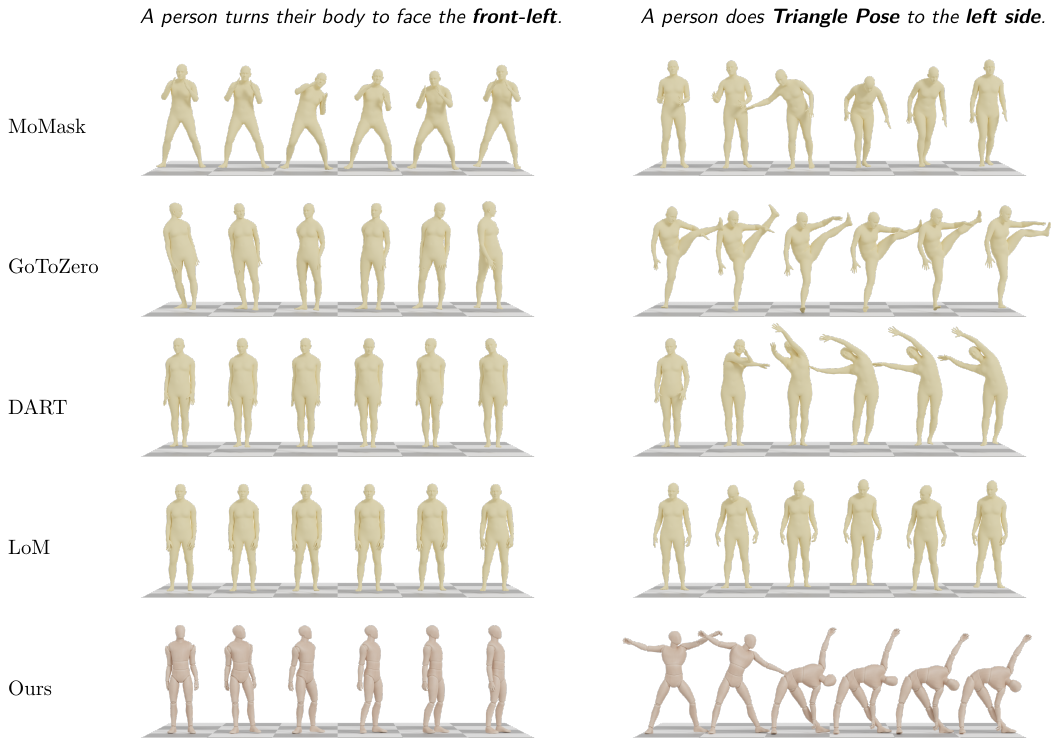} \\ \\
        \includegraphics[width=0.9\linewidth, trim=50 450 45 30]{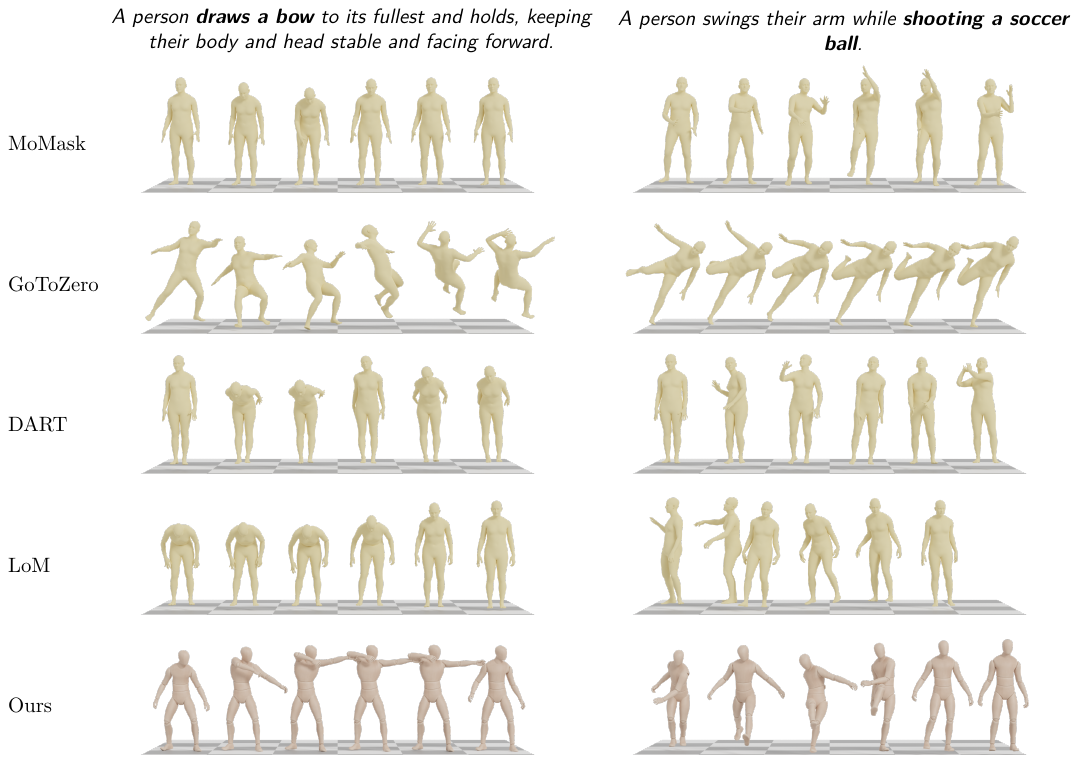}
    \end{tabular}
    \vspace{10pt}
    \caption{
        Examples of visual comparison to state-of-the-art models.
    }
    \label{fig:qualitative}
\end{figure}

To evaluate \modelname{} against state-of-the-art text-to-motion models, we constructed a diverse test set comprising over 2000 text prompts. This test set spans six major categories (\cref{subsec:stats}) and covers simple atomic actions as well as concurrent and sequential action combinations. Human annotators were asked to rate the generated motions on a scale of 1 to 5 from two perspectives: instruction-following capability and motion quality. Fig. \ref{fig:teasercomp} and Tabs. \ref{tab:sota_comparison}–\ref{tab:sota_comparison_mq} show the comparison with four models: DART \cite{DartControl}, LoM \cite{chen2024language}, GoToZero \cite{Fan_2025_ICCV}, and MoMask \cite{momask}. Fig. \cref{fig:qualitative} presents some examples of the visual comparison. Our model significantly outperforms the other models in terms of both instruction-following capability and motion quality.

We further adopt an automatic evaluation approach, namely Structured Semantic Alignment Evaluation (SSAE) \cite{cao2025hunyuanimage}, which utilizes powerful video-VLMs to assess the generated motions. This approach transforms the text-motion alignment problem into a video-question-answering task. It works by decomposing each input prompt into a series of ``yes'' or ``no'' questions. These questions are then presented to a video-VLM (\eg, Gemini-2.5-Pro) along with the rendered video of the generated motion. For example, given the prompt ``a person swings their arm while shooting a soccer ball'', the decomposed questions might include: ``is the person kicking their leg?'', ``is the person swinging their arm?'', and ``does the person appear to be shooting a soccer ball?''. The correct rate of the VLM's answers across all test prompts constitutes the model's SSAE score. A comparison of SSAE scores with other models is provided in \cref{tab:sota_comparison}.

\subsection{Scaling Experiments}
\label{subsec:ablation}

To understand the effects of scale, we investigated how model size and data volume impact performance. The results of our human evaluation on models of varying sizes are presented in Tabs. \ref{tab:scalingexp}–\ref{tab:scalingexp_mq}. Our findings indicate that while instruction-following capability consistently improves with larger models, motion quality reaches a saturation point beyond the 0.46B parameter size. Notably, a comparison between the "DiT-0.46B" and "DiT-0.46B-400h" models (Tab. \ref{tab:scalingexp}) underscores the critical role of the larger 3,000-hour training dataset in enhancing instruction-following capability. Consequently, we plan to release "DiT-1B" as the primary HY-Motion 1.0 model and "DiT-0.46B" as a lightweight "Lite" version.

\begin{table*}[tb]
    \centering
    \caption{\textbf{Instruction-following capability comparison} of different model sizes. Motion categories: (a) Locomotion, (b) Sports \& Athletics, (c) Fitness \& Outdoor Activities, (d) Daily Activities, (e) Social Interactions \& Leisure, and (f) Game Character Actions. The model ``DiT-0.46B-400h'' is trained only on the 400-hour high-quality dataset, while the other models are pretrained on the 3,000-hour dataset.}
    \vspace{5pt}
    \label{tab:scalingexp}
    \begin{tabular}{l|cccccc|c}
        \toprule
        Method & Catg. (a) & (b) & (c) & (d) & (e) & (f) & Avg.   \\
        \midrule
        DiT-0.05B & 3.68 & 2.95 & 2.85 & 3.10 & 3.12 & 2.90 & 3.10  \\
        DiT-0.46B & 3.93 & 3.15 & 3.02 & 3.10 & 3.15 & 2.88 & 3.20 \\
        DiT-0.46B-400h & 3.76 & 2.95 & 2.73 & 2.93 & 3.17 & 2.75 & 3.05 \\
        DiT-1B & \textbf{3.95} & \textbf{3.32} & \textbf{3.23} & \textbf{3.36} & \textbf{3.27} & \textbf{2.92} & \textbf{3.34}  \\
        \bottomrule
    \end{tabular}
    \vspace{-10pt}
\end{table*}

\begin{table*}[tb]
    \centering
    \caption{\textbf{Motion quality comparison} of different model sizes. Motion categories: (a) Locomotion, (b) Sports \& Athletics, (c) Fitness \& Outdoor Activities, (d) Daily Activities, (e) Social Interactions \& Leisure, and (f) Game Character Actions.}
    \vspace{5pt}
    \label{tab:scalingexp_mq}
    \begin{tabular}{l|cccccc|c}
        \toprule
        Method & Catg. (a) & (b) & (c) & (d) & (e) & (f) & Avg.  \\
        \midrule
        DiT-0.05B & 3.32 & 2.54 & 2.95 & 2.86 & 3.00 & 2.80 & 2.91  \\
        DiT-0.46B & 3.68 & \textbf{3.32} & 3.10 & 3.10 & 3.22 & 3.13 & 3.26 \\
        DiT-0.46B-400h & \textbf{3.83} & 3.02 & 3.23 & \textbf{3.29} & 3.27 & 3.20 & 3.31 \\
        DiT-1B & 3.78 & 3.20 & \textbf{3.33} & 3.26 & \textbf{3.27} & \textbf{3.23} & \textbf{3.34}  \\
        \bottomrule
    \end{tabular}
\end{table*}

\section{Conclusion}
\label{sec:conclusion}

In this work, we presented HY-Motion 1.0, a series of large-scale motion generation models that establish a new state-of-the-art in the text-to-motion domain. By successfully scaling a DiT-based flow matching architecture and implementing a meticulous data curation pipeline alongside a comprehensive, full-stage training paradigm, we have provided a clear and effective path for developing high-performance, instruction-following motion generation models. Our work yields key insights into the principles of scaling text-to-motion models:

\begin{itemize}
    \item \textit{The Duality of Data Scale and Quality}: Our experiments reveal a clear distinction in how data properties affect model capabilities. We found that scaling the volume of training data is the primary driver for enhancing instruction following and semantic understanding. In contrast, improving the quality of the data through careful curation is the decisive factor for increasing motion fidelity and physical realism.

    \item \textit{Effectiveness of the Multi-Stage Training Paradigm}: We demonstrate that our three-stage training framework — large-scale pretraining, high-quality fine-tuning, and reinforcement learning — is essential. This "coarse-to-fine" approach effectively balances the trade-off between motion diversity and precision, suggesting that this data-centric, multi-stage optimization strategy is a robust path forward for the field.
\end{itemize}

\paragraph{Limitations.}
Despite these advancements, HY-Motion 1.0 exhibits certain limitations that point towards future research directions:

\begin{itemize}
    \item \textit{Complex Instructions}: While our model significantly outperforms baselines in semantic alignment, it still faces challenges with highly detailed or complex instructions. This is partially due to the inherent difficulty of our data captioning pipeline, i.e., creating complete and accurate textual descriptions for nuanced and intricate motions remains a significant challenge for both VLM-based captioning and the manual refinement process. 
    
    \item \textit{Human-Object Interaction (HOI)}: As our current dataset primarily focuses on body kinematics without explicit object geometry, the model may struggle to generate physically accurate interactions with external objects (e.g., precise contact points when holding a tool).
\end{itemize}

We are releasing HY-Motion 1.0 to the open-source community and hope it serves as a solid baseline, inspiring further exploration and accelerating the development of scalable, high-quality motion generation technologies.

\newpage
\section*{Contributors} %
\label{sec:contributors}

\begin{itemize}[leftmargin=0.25cm]
    \item \textbf{Project Sponsors:} Jie Jiang, Linus, Yuhong Liu
    \item \textbf{Project Supervisor:} Xiaolong Li
    \item \textbf{Project Leader:} Linchao Bao
    \item \textbf{Core Contributors:} Yuxin Wen, Qing Shuai, Di Kang, Jing Li, Cheng Wen, Yue Qian, Ningxin Jiao, Changhai Chen
    \item \textbf{Contributors:} 
    
    \begin{itemize}[leftmargin=0.5cm]
        \item \textbf{Product/Project Managers:} Weijie Chen, Yiran Wang, Jinkun Guo, Dongyue An
        \item \textbf{Artists:} Han Liu, Yanyu Tong, Chao Zhang, Qing Guo, Juan Chen, Qiao Zhang
        \item \textbf{Engineering:} Youyi Zhang, Zihao Yao, Cheng Zhang, Hong Duan, Xiaoping Wu, Qi Chen
        \item \textbf{Data:} Fei Cheng, Liang Dong, Peng He, Hao Zhang, Jiaxin Lin, Chao Zhang
        \item \textbf{Evaluation:} Zhongyi Fan, Yifan Li, Zhichao Hu
    \end{itemize}
\end{itemize}

\clearpage

{\small
\bibliographystyle{plain}
\bibliography{references}
}

\end{document}